\documentclass{article}
\PassOptionsToPackage{numbers,compress}{natbib}
\usepackage[final]{neurips_2026}
\makeatletter
\renewcommand{\@noticestring}{Workshop on Who Verifies the Agents? Toward Reliable Agent Development, at 40th Conference on Neural Information Processing Systems (NeurIPS 2026).}
\makeatother

\usepackage[utf8]{inputenc}
\usepackage[T1]{fontenc}
\usepackage{microtype}
\usepackage{booktabs}
\usepackage{array}
\usepackage{pifont}
\usepackage{threeparttable}
\usepackage{amsmath,amssymb}
\usepackage{xcolor}
\usepackage{tikz}
\usetikzlibrary{positioning,arrows.meta,fit,backgrounds,calc,shapes.geometric}
\usepackage{graphicx}
\usepackage{caption}
\usepackage{needspace}
\usepackage{placeins}
\usepackage[hidelinks]{hyperref}
\usepackage{url}

\definecolor{detfg}{HTML}{009E73}\definecolor{detbg}{HTML}{D6F0E6}
\definecolor{learnfg}{HTML}{0072B2}\definecolor{learnbg}{HTML}{D5E7F5}
\definecolor{hybfg}{HTML}{E69F00}\definecolor{hybbg}{HTML}{FBEBD0}
\definecolor{waistfg}{HTML}{4053A3}\definecolor{waistbg}{HTML}{E6E9F5}

\title{FinVerify: A Verification Framework for Financial Agents}

\author{%
  Mao Guan$^{1}$, \quad Qian Chen$^{2}$, \quad Ge Meng$^{3}$, \quad Xiao-Yang Liu$^{4}$\\
  $^{1,2}$Independent Researcher \quad $^{3}$New York University
  \quad $^{4}$Columbia University\\
  Email: \texttt{XL2427@columbia.edu}
}

\begin{document}
\maketitle
 
\begin{abstract}
Financial agents report numerical results critical to investment analysis and decision-making. Their outputs may contain errors involving reporting periods, units, and accounting relationships that general-purpose claim verifiers do not explicitly check. A large language model (LLM) judge's verdict alone does not establish whether these requirements have been checked. In this paper, we propose \textit{FinVerify}, a financial claim-verification framework motivated by the Proposal--Certification--Execution (PCE) architecture. From 190 incorrect trajectories in the Agent Finance Reasoning dataset, we derive a taxonomy of five financial error categories: Temporal Integrity, Source Fidelity, Comparability, Accounting Integrity, and Financial-Model Fidelity. FinVerify operationalizes each category as a verification protocol: an evidence-bound LLM structures the inputs, and deterministic validators check the applicable requirements. Where applicable, Z3 checks whether the grounded inputs and an authoritative formula are consistent with the reported value. FinVerify returns Reject when a violation is established and Abstain when required evidence is insufficient; acceptance requires every applicable protocol to pass. We evaluate FinVerify against a Direct baseline and an LLM judge on five benchmarks, using the same fixed candidate sets across methods. On Agent Finance Reasoning, FinVerify increases accepted-claim precision from the LLM judge's 56.4\% to 76.7\%, with 81.2\% coverage. On XBRLFiling, FinVerify checks each answer against facts and calculation relationships filed with the United States Securities and Exchange Commission (SEC), achieving 99.8\% precision and 97.0\% decision accuracy, compared with the judge's 58.0\% decision accuracy. On FinanceBench-67, FinVerify achieves 87.9\% precision, compared with the judge's 69.7\%; the four incorrect answers it accepts contain values that appear in, or can be derived from, the benchmark's free-text reference answers. On Financial Question Answering (FinQA) and the Tabular and Textual Dataset for Question Answering (TAT-QA), where the available evidence consists only of a passage and a table, the precision differences are smaller. The judge tends to reject correct answers across the benchmarks, whereas FinVerify achieves higher decision accuracy on non-abstained candidates across all five, at the cost of more abstentions.
\end{abstract}

\section{Introduction}
\label{sec:intro}
Financial agents powered by large language models
(LLMs)~\cite{fingpt_data,fingpt,finagent,tradingagents,finmem} report numerical results that are critical to investment analysis and decision-making. They retrieve filings, extract figures,
compute ratios and growth rates, and provide answers with citations.
Their outputs inform investment and reporting decisions in a regulated domain.
Recent studies make the verification burden concrete. Deep-research agents screen reverse-merger candidates for mergers and acquisitions (M\&A) advisors, yet checking their cited evidence remains time-consuming~\cite{shellscreening}. In LLM-generated investment research memos, low claim-level error rates can coexist with substantial memo-level exposure to claims contradicted by official disclosures~\cite{investmentmemos}.

Re-executing an agent’s program can reproduce its calculation, but does not establish whether the inputs and formula are appropriate for the question. For example, a calculation using revenue from the wrong reporting period may run without error and still produce an incorrect answer. NumTrace~\cite{numtrace} documents ``right number, wrong account'' (RNWA) errors: a claim can pass a value check while failing to match the correct entity, concept, period, unit, or scope. Together, these studies motivate independent checks of the evidence and financial meaning behind an agent's claims.
General-purpose claim verifiers~\cite{factscore,factool,safe} check text
against retrieved passages but do not explicitly check reporting periods, units, accounting relationships, or canonical financial formulas. A cited passage may contain the reported number without establishing whether that number is appropriate for the question. Finance-specific
methods~\cite{verifin,finground,finverbench} address several of these
requirements, motivating a common framework for organizing the checks.
Self-critique~\cite{critic,huang2024} relies on the model under test, and LLM
judges~\cite{judges} return a verdict without a machine-checkable account of
which verification checks failed.

The Proposal--Certification--Execution (PCE) architecture of Yanglet et~al.~\cite{certifiedtraces} separates agent proposal, independent certification, and execution authorization. Generating a plausible output does not itself authorize its use in a downstream action. For financial agents, this motivates a concrete verification problem: which financial requirements must a claim and its recorded derivation satisfy, and what evidence establishes each requirement? FinVerify addresses this financial claim-verification component through explicit protocols and evidence-backed decisions. Full trace certification and execution enforcement remain beyond this study's scope.

\paragraph{How financial agents fail.}
We examine all 190 trajectories labeled incorrect in the Snorkel Agent Finance Reasoning
dataset~\cite{snorkel}, using each trajectory's recorded tool calls and the dataset's
correctness rationale. Asked which of
Walmart and Tesla has the lower concentration ratio, an agent compared
Walmart's fiscal-2025 figure with Tesla's fiscal-2024 figure; using data not yet available at the decision date introduces look-ahead bias
into forecasting benchmarks~\cite{guan2026leakage}. Asked how
Walmart's domestic and international operations performed, an agent queried
the total income-before-tax table instead of income from continuing
operations and reported that income outside the United States had fallen 8.9\%, when it had actually risen 65.6\%.
Asked how significant Caterpillar's Resource Industries segment is, an agent
divided segment revenue by total company revenue, obtaining 19.1\%, rather than by
reportable-segment revenue, which would yield 17.5\%, although it had retrieved both
denominators. Asked how Citibank's pension expense responds to lower rates, an
agent computed $-27+21=-6$ and then reported a \$6 million \emph{increase}. Asked
about Apple's lease maturity profile, an agent reported that 43.3\% of
obligations fall beyond five years, called this the majority, and concluded
that the schedule ``minimizes near-term liquidity pressure.''

\paragraph{A financial error taxonomy.}
\begin{table}[t]
\centering\small
\caption{Agent errors in the Snorkel Agent Finance Reasoning dataset, classified from the recorded trajectories and the dataset's correctness rationale. Every trajectory the dataset labels wrong receives exactly one primary category.}
\label{tab:tB}\label{tab:errors}
\begin{tabular}{@{}lp{0.48\linewidth}rr@{}}
\toprule
Category & Agent error & Count & Share \\
\midrule
A: Temporal integrity & wrong, stale, future, or unavailable period & 3 & 1.6\% \\
B: Source fidelity & missing, wrong, or unsupported evidence & 145 & 76.3\% \\
C: Comparability & incompatible entity, unit, metric, or basis & 8 & 4.2\% \\
D: Accounting integrity & arithmetic, sign, or accounting-identity error & 2 & 1.1\% \\
E: Financial-model fidelity & wrong definition, method, formula, or interpretation & 32 & 16.8\% \\
\midrule
\textbf{All agent errors} & Trajectories labeled wrong by the dataset & \textbf{190} & \textbf{100.0\%} \\
\bottomrule
\end{tabular}
\end{table}

These failures fall into five categories, which we call
\emph{Temporal Integrity} (the wrong, stale, or not-yet-available period),
\emph{Source Fidelity} (a value or field absent from the
cited source or inconsistent with it), \emph{Comparability} (operands that differ in period, entity,
unit, scale, or basis), \emph{Accounting Integrity} (arithmetic, sign, or
identity errors given the right inputs and formula), and
\emph{Financial-Model Fidelity} (the wrong metric definition, formula, or
interpretation). Table~\ref{tab:errors} assigns every one of the 190 errors to
exactly one category: source fidelity dominates (76\%), financial-model
fidelity is second (17\%), and the remaining three together account for 7\%. The
five categories mirror the assertions auditors
test~\cite{as1105,isa315,isa500,isa520,sfac8,dqc} (cut-off, occurrence,
comparability, accuracy, and classification), so we expect these categories to apply
beyond this dataset. Each is \emph{checkable}: filing timestamps,
facts filed in eXtensible Business Reporting Language (XBRL), typed units
and periods, calculation linkbases, and
canonical metric definitions are objective oracles that open-domain question
answering lacks.

\paragraph{FinVerify.}
In this paper, we propose \textit{FinVerify}, a framework that determines which financial checks apply to an agent's output and combines their outcomes into a final verification decision. Building on source-grounded numerical verification, FinVerify organizes these checks into five protocols with explicit applicability conditions. A successful numerical check cannot override a temporal violation or missing evidence required by another protocol. An LLM extracts the requested metric, numerical inputs, and their financial
context from the question, fixed evidence, and agent trace for subsequent checks. Deterministic
validators drop any quote that does not appear in the evidence. The five protocols apply explicit financial rules. Where an authoritative numerical relation is available, Z3~\cite{z3}, a satisfiability modulo theories (SMT) solver, checks whether the grounded inputs, prescribed formula, and reported answer are mutually consistent; conflicting constraints help explain a rejection. The XBRLFiling adapter uses direct recomputation instead of Z3 (Section~\ref{sec:verifiers}). The validators return Reject when a violation is proved and Abstain when the evidence is insufficient. Acceptance requires every applicable protocol to pass, and the verdict records the applied checks, their supporting evidence, and any failures in a protocol$\to$policy$\to$function trace (Figure~\ref{fig:framework}).

\paragraph{Evaluation.}
We evaluate FinVerify under an Accept/Reject/Abstain protocol on
five benchmarks (Agent Finance Reasoning, FinQA~\cite{finqa},
TAT-QA~\cite{tatqa}, FinanceBench-67~\cite{financebench}, and
XBRLFiling) against a Direct baseline and an LLM judge, using the same fixed candidate sets across methods. On Agent Finance
Reasoning, FinVerify also reads the filing tables behind the recorded tools
(Section~\ref{sec:setup}). FinVerify raises accepted-claim
precision over Direct on every benchmark and over the LLM judge on four of
five. On XBRLFiling, FinVerify checks each answer against facts filed with the United States Securities and Exchange Commission (SEC) and calculation relationships, achieving 99.8\% precision and 97.0\% decision accuracy, compared with the judge's 58.0\% decision accuracy; it accepts one incorrect answer among 596 non-abstaining decisions. On Agent Finance Reasoning, it increases accepted-claim precision from
46.8\% (Direct) and 56.4\% (LLM judge) to 76.7\%, with 81.2\% coverage. On FinanceBench-67, precision is 87.9\%, compared with the judge's 69.7\%. On FinQA and TAT-QA, where the available evidence consists only of a passage and a table, the precision differences are smaller. FinVerify achieves higher decision accuracy than the judge on all five benchmarks, at the cost of more abstentions.

\paragraph{Organization.} Section~\ref{sec:taxonomy} presents the error study of 190 trajectories and the taxonomy; Section~\ref{sec:framework} describes the verification protocols and proof gate. Sections~\ref{sec:setup}--\ref{sec:results} report the experimental setup and results on five benchmarks.

\section{Related Work}
\label{sec:related}
\paragraph{Financial agents and benchmarks.}
Financial language models~\cite{fingpt_data,fingpt} support agents with
multimodal~\cite{finagent}, multi-agent~\cite{tradingagents}, and
memory-augmented~\cite{finmem} designs. These systems analyze filings or market
data and produce numerical answers or investment decisions, motivating
independent checks of their outputs. Broader benchmark suites include FinBen~\cite{finben}, which covers diverse financial tasks, and MultiFinBen~\cite{multifinben}, which evaluates multilingual and multimodal financial tasks using difficulty-aware dataset selection. Benchmarks such as FinQA~\cite{finqa},
TAT-QA~\cite{tatqa}, FinanceBench~\cite{financebench}, and agentic sets such
as Agent Finance Reasoning primarily evaluate final answers against references.
The error study in Section~\ref{sec:taxonomy} complements these evaluations by
classifying the failures observed in financial-agent trajectories.

\paragraph{General-purpose verification.}
Atomic-claim verifiers~\cite{factscore,factool,safe} decompose long-form text
into statements and assess their support against evidence; LLM-as-judge
methods~\cite{judges} use a model to assess candidate outputs. These approaches
can assess factual support, but do not necessarily provide explicit checks
of reporting periods, units, or accounting identities
(Table~\ref{tab:related}, appendix). Self-correction without an external signal
can be unreliable~\cite{huang2024}, motivating tool-augmented
critique~\cite{critic}. FinVerify provides such an external signal through financial checks with
recorded evidence and outcomes.

\paragraph{Finance-specific verification.}
VeriFin~\cite{verifin} grounds numerical answers in filed XBRL facts and
binds reporting periods and units. Z3 checks whether bound values, documented
formulas, and reported answers are consistent, and identifies conflicting
constraints through unsatisfiable (UNSAT) cores. FinVerify's source and accounting protocols follow this
grounding-and-recomputation pattern; we also evaluate FinVerify on two of
VeriFin's candidate sets. FinGround~\cite{finground} decomposes text into numerical,
temporal, comparative, and computational claims, grounds them in retrieved
passages, and recomputes computational claims from formula templates.
AuditFlow~\cite{auditflow} separates agent-guided search from deterministic,
taxonomy-based verification of XBRL filings.
FinVerBench~\cite{finverbench} studies benchmark validity for statement
verification and is not itself a verifier. NumTrace~\cite{numtrace} is a
benchmark that scores the value and the binding (entity, concept, period,
unit, scope) of each numerical claim separately; most failing claims of the
agents it audits pass the value check but fail the binding check. Lin and
Liu~\cite{shellscreening} specify, but do not evaluate, a certification layer
for deep-research screening reports. Decision-time leakage and
look-ahead bias have dedicated
benchmarks~\cite{leakage,lookahead,exante,guan2026leakage}; these benchmarks
isolate temporal integrity (A), but as a property of a backtest or a
point-in-time model rather than of a cited claim. Existing verifiers also
cover aspects of comparability (C) and financial-model fidelity (E):
VeriFin binds periods and units on XBRL facts, and FinGround reconstructs
formulas from text. FinVerify organizes these checks into five protocols
derived from an empirical study of financial-agent failures and composes
them behind one verification interface.

\paragraph{Audit assertions.}
Auditing standards~\cite{as1105,isa315,isa500} decompose ``is this statement
right?'' into assertions (occurrence, completeness, accuracy, cut-off,
classification, presentation), each testable against independent evidence;
analytical procedures~\cite{isa520} add reasonableness, the Financial Accounting Standards Board (FASB) conceptual
framework~\cite{sfac8} adds comparability, and the XBRL US Data Quality Committee (DQC)
ruleset~\cite{dqc} makes identities machine-checkable. This decomposition motivates checking distinct financial properties against
independent evidence; Section~\ref{sec:taxonomy} connects it to the observed
error categories.

\section{Method}
\label{sec:framework}
FinVerify is a verification framework for financial agents. This section describes the method and design of FinVerify, including the problem formulation (\S\ref{sec:problem}), empirical
error study (\S\ref{sec:taxonomy}), and FinVerify implementation details
(\S\ref{sec:repr}--\S\ref{sec:adapters}).

\subsection{Problem setting}
\label{sec:problem}
A verification instance consists of a question $q$, frozen evidence $E$
(filing text, tables, filed XBRL facts, and an optional decision-time
\emph{as-of} date), and a claim $C=\langle v, P, \mathcal{C}\rangle$ exposed
by an agent: the reported value $v$, the program $P$ that produced it, and the
citations $\mathcal{C}$ it gives. The verifier returns one of
\textsc{Accept}, \textsc{Reject}, or \textsc{Abstain}, together with a proof
trace naming the protocol, policy, function, and evidence span behind the
decision. \textsc{Accept} means that the claim $C$ is supported by the evidence $E$ and by
independent authorities on every applicable protocol; \textsc{Reject} means
a violation has been \emph{proved}; \textsc{Abstain} means the required proof
could not be established.

\subsection{From agent errors to protocols}
\label{sec:taxonomy}
We derive FinVerify's five verification protocols from an error analysis of the Snorkel Agent Finance Reasoning dataset~\cite{snorkel}. We examine all 190 trajectories labeled incorrect and use a single LLM call to classify each based on its recorded tool calls and the dataset's correctness rationale. Each trajectory is assigned one primary error category, defined by the earliest cause of the incorrect answer. The classification is independent of FinVerify's outputs, and the resulting labels are not human-validated (see Section~\ref{sec:limitations}).

Table~\ref{tab:errors} reports the distribution across the five categories introduced in Section~\ref{sec:intro}. \textbf{B~Source Fidelity} accounts for 76\% of errors, indicating that unsupported or incorrectly sourced values are the dominant failure mode. In contrast, only two errors fall under \textbf{D~Accounting Integrity}, suggesting that re-executing an agent's calculations alone would detect only a small fraction of failures. These findings motivate five protocols, each grounded in an independent financial authority: filing dates for \textbf{A~Temporal Integrity}, cited documents and filed XBRL facts for \textbf{B~Source Fidelity}, recorded periods, units, and entities for \textbf{C~Comparability}, program re-execution and issuer-filed calculation identities for \textbf{D~Accounting Integrity}, and canonical metric definitions for \textbf{E~Financial-Model Fidelity}. Together, these protocols verify not only the calculations but also the evidence, context, and financial definitions underlying each claim.

\subsection{Shared representation and catalog}
\label{sec:repr}
\begin{figure}[t]
\centering
\includegraphics[width=0.87\linewidth]{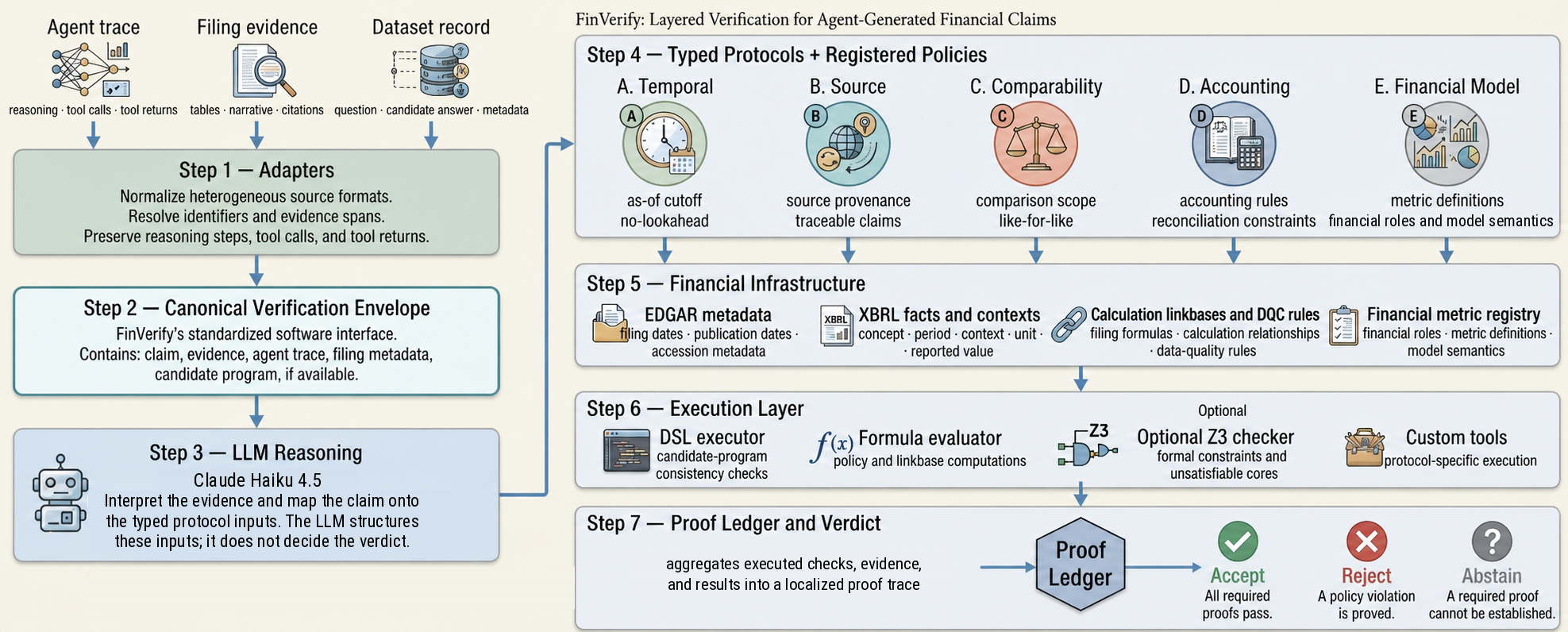}
\caption{In FinVerify, an LLM proposes typed bindings while fixed rules determine which protocols apply, and the proof gate accepts only if all required protocols pass, rejects proven policy violations, and abstains otherwise.}
\label{fig:framework}
\end{figure}
\begin{table}[t]
\centering
\footnotesize
\caption{ The executable catalog includes five protocols, nine versioned policies, and twenty registered functions, each relying on candidate-independent sources of authority; function names are domain-specific, so the accounting and financial-model \texttt{solve\_formula} functions are distinct, and the LLM can only reference existing functions, not add or remove them. Abbreviations: Financial Statement Data Sets (FSDS), Electronic Data Gathering, Analysis, and Retrieval (EDGAR), and domain-specific language (DSL).}
\label{tab:catalog}
\begin{tabular}{@{}>{\raggedright\arraybackslash}p{3.65cm}>{\raggedright\arraybackslash}p{5.25cm}>{\raggedright\arraybackslash}p{3.95cm}@{}}
\toprule
Protocol, policy & Functions & Authority \\
\midrule
A \texttt{no\_lookahead.v1} & resolve\_as\_of, resolve\_publication, compare\_cutoff & question; SEC FSDS/EDGAR; date order \\
B \texttt{traceable\_claim.v1} & resolve\_citation, ground\_operands, match\_filed\_value & frozen evidence; filed XBRL \\
C \texttt{like\_for\_like.v1} & compare\_period, compare\_unit, compare\_scope & typed XBRL contexts \\
C \texttt{requested\_unit.v1} & compare\_requested\_unit & question; typed operand scale \\
C \texttt{percent\_scale.v1} & compare\_percent\_scale & re-executed program; question \\
D \texttt{claim\_consistency.v1} & execute\_program, check\_identities & deterministic DSL; filed facts \\
D \texttt{authoritative\_}\newline\texttt{formula.v1} & load\_linkbase, solve\_formula & calculation linkbase; Z3 \\
E \texttt{authorized\_}\newline\texttt{definition.v1} & resolve\_metric, bind\_roles, solve\_formula, check\_bounds & metric registry; filed facts; Z3 \\
E \texttt{question\_stated\_}\newline\texttt{definition.v1} & apply\_question\_definition & definition stated in the question \\
\bottomrule
\end{tabular}
\end{table}

Every instance is verified through one representation, $\langle$Filing,
Claim, VerificationPlan, Policy, FunctionResult, Verdict$\rangle$
(Figure~\ref{fig:framework}). A \emph{Filing} holds text, tables, and filed
XBRL facts. A \emph{Claim} holds $v$, $P$, and $\mathcal{C}$ together with
the typed operands bound by fitting (\S\ref{sec:fit}). A
\emph{VerificationPlan} lists which protocols are required. A versioned
\emph{Policy} states its applicability, the functions it may call, and its
missing-evidence behaviour. Every \emph{FunctionResult} records a status in
$\{\textsc{pass},\textsc{fail},\textsc{abstain}\}$, a failure code, the
authority consulted, and the evidence spans involved; the \emph{Verdict}
retains the full protocol$\to$policy$\to$function trace.
The executable contract is a three-level catalog (Table~\ref{tab:catalog}):
five protocols, nine versioned policies, and twenty registered functions,
each tagged with the authority it consults.

\subsection{Evidence-bound fitting}
\label{sec:fit}
On the standard pathway, a single language-model call converts the question, evidence, and agent trace into a structured verification record. The record identifies any stated \emph{as-of} date, the requested metric, the agent's computation, and the source document for each citation. For each numerical operand, it also records a verbatim quote from the evidence and the associated period, unit, scale, entity, and accounting basis. A deterministic validator then checks whether each quote appears in the evidence and contains the number it is used to support. If either check fails, the quote is discarded and the operand remains untyped. Since untyped operands cannot satisfy source-grounding requirements, the verifier returns \textsc{Abstain} rather than \textsc{Accept} when the required evidence is missing. This validation prevents the language model from establishing support through fabricated quotes, but it does not guarantee that the extracted metadata are correct. Missing operands or discarded quotes reduce coverage, while incorrectly assigned periods or units may affect subsequent checks. We do not measure the latter effect separately. For the more complex, multi-part trajectories in Agent Finance Reasoning, the trace-native front end described in Section ~\ref{sec:verifiers} uses several language-model calls to prepare the verification inputs. In both pathways, the deterministic protocols and proof gate, rather than the language model, determine the final verdict.

\subsection{Validated planning}
\label{sec:plan}
The plan is built from the validated fit by fixed rules, not chosen by the
model. \textbf{B}~is required for every numeric claim. \textbf{A}~is
required when an as-of cutoff resolved. \textbf{D}~is required when the program
contains an operator or the issuer declares a
calculation identity for the claimed concept.
\textbf{C}~is required when at least two typed operands share a dimension that
the dataset policy marks as required, or the question
names the unit the answer must use, or the stated value
and the candidate's own re-executed program differ by a percent-to-ratio
rescale. \textbf{E}~is required when the metric resolves
to a registry definition or the question states
the definition the derivation must follow.
Each policy runs only when its applicability conditions are met. It returns \textsc{Pass} when the check succeeds, \textsc{Fail} when a violation is proved, and \textsc{Abstain} when the evidence is insufficient. Policies that do not apply are excluded from the verification plan.

\subsection{The five protocols}
\label{sec:protocols}
Each protocol applies its policies through the registered functions in the executable catalog. \textbf{B~Source Fidelity} provides the evidence bindings used by the other four protocols, making source grounding the foundation of the verification process.

\begin{itemize}\setlength\itemsep{1pt}

\item \textbf{A~Temporal Integrity} (\texttt{no\_lookahead}). This protocol checks whether the cited information was publicly available by the specified \emph{as-of} date. For each cited document, FinVerify resolves its accession number against a frozen registry of SEC Financial Statement Data Sets (FSDS) and the Electronic Data Gathering, Analysis, and Retrieval (EDGAR) system and checks that $\mathrm{available}(d)\le\mathrm{as\_of}$ for all $d\in\mathcal{C}$. A document published after the cutoff constitutes a violation, while an unresolved publication date leads to \textsc{Abstain}. On FinanceBench-67 and XBRLFiling, publication dates are obtained from SEC filing records; on other benchmarks, the protocol uses the dates attached to citations. 

\item \textbf{B~Source Fidelity} (\texttt{traceable\_claim}). This protocol checks whether the numbers used in a claim are supported by the evidence. It locates the cited passages in $E$, checks whether each number used in the calculation appears in the supporting evidence, and compares reported values with filed XBRL facts when available. If a reported value contradicts a filed fact, the claim fails. If the evidence is missing or cannot be located, the protocol returns \textsc{Abstain}, since a lack of evidence does not prove that a value is wrong. When the agent provides no citations, FinVerify uses validated quotes from the available evidence or matches the reported value to the relevant filed XBRL fact identified by the fitter.

\item \textbf{C~Comparability} (\texttt{like\_for\_like}, \texttt{requested\_unit}, \texttt{percent\_scale}). This protocol checks whether the numbers used in a calculation are comparable and whether the final answer uses the unit and scale requested in the question. It compares each operand's reporting period, unit, scale, entity or segment, and accounting basis. A mismatch in any required dimension is a violation. It also checks whether the reported value follows the question's unit and scale requirements. For example, if the question asks for billions of United States dollars (USD) but the answer uses the filing's original units, the claim fails. Similarly, multiplying a result by 100 when the question does not ask for a percentage is treated as a scale error. This error is checked here rather than by the accounting protocol (Section~\ref{sec:verifiers}).

\item \textbf{D~Accounting Integrity} (\texttt{claim\_consistency}, \texttt{authoritative\_formula}). This protocol checks whether the agent's calculations are correct and follow established accounting rules. It first re-executes the agent's program and checks whether the results satisfy the relevant accounting identities. When an issuer's calculation linkbase or a DQC rule~\cite{dqc} provides an authoritative formula, FinVerify also checks the reported value against that formula using Z3. This additional check applies only when a valid accounting identity is available. On XBRLFiling, 591 of the 600 candidates meet this condition and are verified through the recomputation procedure in Section~\ref{sec:verifiers}, without Z3. For the remaining nine candidates, FinVerify checks only whether the agent's calculation is internally consistent. The verifier can still make a decision in these cases, and five of the nine candidates are accepted.

\item \textbf{E~Financial-Model Fidelity} (\texttt{authorized\_definition}, \texttt{question\_stated\_definition}).

This protocol checks whether the agent uses the correct financial formula to answer the question. FinVerify first identifies the requested metric in a versioned registry, matches the required inputs to filed financial data, and checks the calculation against the standard formula, including its sign, scale, and permitted bounds. A calculation that conflicts with the applicable definition is a violation. If no applicable definition is available, the policy is not selected. If a definition applies but its required inputs cannot be verified, the policy returns \textsc{Abstain}. If the question provides its own formula, that formula takes priority over the registry. For example, using fiscal year (FY) and property, plant, and equipment (PP\&E) notation, if a question defines a ratio as ``FY2019 revenue / (average PP\&E between FY2018 and FY2019)'', FinVerify should not use the registry's general \texttt{revenue / assets} formula. Instead, it checks whether the agent's program calculates the required average over the two fiscal years. 
\end{itemize}

Protocols D and E share an authority engine that verifies numerical claims against filed facts and established financial relations. Re-executing the agent's program establishes internal consistency but cannot determine whether the correct formula or inputs were used. Where a complete authoritative relation is available, FinVerify independently computes the expected value and checks the constraint

\[
\Phi=\bigwedge_i (x_i=\mathrm{filed}_i)\;\wedge\;(\hat v=f(x))\;\wedge\;(|\hat v-v|\le\tau)
\]

using Z3~\cite{z3}, where $\tau$ is a relative tolerance scaled to the magnitudes involved. If the constraints are unsatisfiable, Z3 returns an \textsc{unsat} core identifying the conflicting operand, formula, or reported value. This provides evidence for the violation. When the required relation is incomplete or unavailable, the engine abstains. Together with the operand grounding provided by Protocol B, this forms the authoritative numerical verification approach of~\cite{verifin}, incorporated here as one component of FinVerify's five-protocol framework.

\subsection{The proof gate}
\label{sec:gate}
The proof gate determines the final verdict from the results of the required protocols, without relying on the language model's judgment. It returns \textsc{Accept} only when every required protocol passes. If any required protocol proves a violation, the gate returns \textsc{Reject}, together with the supporting evidence and the corresponding protocol$\to$policy$\to$function trace. If no violation is proved but a required protocol cannot complete its check, the gate returns \textsc{Abstain}. It also abstains when no protocol is applicable. Only authoritative checks can establish a violation. A failure reported by a heuristic function, such as a learned entailment model, is treated as \textsc{Abstain} rather than \textsc{Reject}. Policies that do not apply to a claim are excluded from verification, whereas applicable policies that cannot establish a result lead to abstention. Thus, a successful check cannot override missing evidence or a proven violation in another protocol, and every rejection must be supported by a deterministic check.

\subsection{Adapters}
\label{sec:adapters}
FinVerify uses dataset and agent adapters to support different benchmarks and agent outputs without changing the verification framework. The dataset adapter converts each benchmark's evidence into a standard \emph{Filing} representation and provides the information needed to apply the policies, including any \emph{as-of} date, required comparability dimensions, and available accounting identities. We implement dataset adapters for all five benchmarks. The agent adapter converts the agent's output into a \emph{Claim} containing its reported answer, program, and citations. FinQA and TAT-QA use this standard pathway, while the other benchmarks use the specialized front ends described in Section~\ref{sec:verifiers}.

\section{Experimental Setup}
\label{sec:setup}
Every verifier judges the same fixed candidates over frozen evidence;
candidates are generated once and never regenerated, controlling for
candidate quality.
Table~\ref{tab:pools} (appendix) summarizes the five benchmarks and marks the
authoritative sources that FinVerify alone reads.

\subsection{Benchmarks and candidate sets}
\label{sec:pools}
A verification benchmark needs candidates labeled correct or wrong. FinQA
and TAT-QA ship gold answers, not candidates, so we construct the candidates
and derive the labels from the gold. For FinanceBench-67 and XBRLFiling we use
the candidates and labels released with VeriFin~\cite{verifin}, which were
also generated by Claude Haiku~4.5.

\paragraph{Question answering (QA) benchmarks.}
FinQA~\cite{finqa} and TAT-QA~\cite{tatqa} contain numerical questions based on financial tables and text. We sample questions with numerical reference answers from their development sets using a fixed seed, restricting TAT-QA to arithmetic questions. FinanceBench-67~\cite{financebench} uses annotated passages from SEC filings, mainly annual (Form~10-K) and quarterly (Form~10-Q) reports. XBRLFiling asks for financial statement subtotals from United States companies, with the target values removed but their components retained. The original data set values serve as reference answers. For FinQA and TAT-QA, Claude Haiku~4.5 generates candidate answers from fixed evidence. We extract one numerical claim per answer, normalize its unit and label it as Correct if it falls within 1\% or one unit of the reference value. For FinanceBench-67 and XBRLFiling, we matches candidate values against numbers in the reference text, including manual adjudication for some FinanceBench questions. 
\paragraph{SEC filings.}
For FinanceBench-67 and XBRLFiling, FinVerify retrieves the underlying SEC filings and uses their EDGAR accession numbers to obtain fixed snapshots of filed XBRL facts and calculation linkbases. When XBRL data is unavailable, verification relies on the available evidence passages. 

\paragraph{Agent Finance Reasoning.}
We use the Snorkel Agent Finance Reasoning dataset~\cite{snorkel} directly, as described in Section~\ref{sec:taxonomy}. Each of its 357 agent trajectories serves as a candidate, with the final answer treated as the claim and the dataset's correctness label as ground truth. Of these, 190 are labeled incorrect. The trajectories contain multi-step financial analyses from five backbone models. Verifiers receive the question and full agent trajectory, including tool outputs, but do not have access to the reference answer, correctness label, or rationale.

\subsection{Verifiers and metrics}
\label{sec:verifiers}
\paragraph{Baselines.}
We compare FinVerify against two baselines. The \emph{Direct} baseline accepts any candidate that provides an answer and abstains otherwise, representing a workflow without verification. The \emph{LLM judge} evaluates each candidate using a zero-shot prompt containing the question, frozen evidence, and candidate answer. For the four QA and filing benchmarks, we use the evidence-only judging prompt from VeriFin~\cite{verifin}. For Agent Finance Reasoning, the judge also reads the full agent trajectory. It returns \textsc{Accept} or \textsc{Reject}, with one retry if the response cannot be parsed, and abstains if no claim is present or the retry fails.  

\paragraph{FinVerify.}
FinVerify uses different verification pathways depending on the benchmark. For FinQA and TAT-QA, the layered engine in Section~\ref{sec:framework} checks whether the numbers are supported by the provided text and tables, whether the operands are comparable, and whether the calculation produces the reported answer using the correct financial definition. For FinanceBench-67 and XBRLFiling, candidates contain only a numerical answer, without a program or citations. The fitter reconstructs the calculation from the evidence and identifies the relevant financial concept. FinVerify then checks the answer against the supporting passages, filed XBRL facts, and applicable accounting identities.

For Agent Finance Reasoning, which contains multi-part analytical answers, FinVerify uses a separate front end. Multiple language-model calls identify the facts required by the question, extract the agent's claims, and match them to the underlying 10-K tables. The same deterministic protocols and proof gate then evaluate these claims. FinVerify also checks whether the answer covers the claims required by the question. A missing required claim is treated as a completeness violation, while a claim that is present but cannot be supported by the available evidence leads to \textsc{Abstain}.
\paragraph{Backbones.}
Every language-model step uses Claude Haiku~4.5: FinVerify's fitter and its
Agent Finance Reasoning front end, the LLM judge, the QA candidate generator,
and the taxonomy labels. 

\paragraph{Metrics.}
We evaluate each verifier by comparing its \textsc{Accept}, \textsc{Reject}, or \textsc{Abstain} decision against the candidate's correctness label. This gives five outcomes: true accepts (TA), false accepts (FA), true rejects (TR), false rejects (FR), and abstentions (Abs). We report the number of false accepts, accepted-claim precision $\mathrm{TA}/(\mathrm{TA}+\mathrm{FA})$, decision accuracy $(\mathrm{TA}+\mathrm{TR})/(\mathrm{TA}+\mathrm{FA}+\mathrm{TR}+\mathrm{FR})$, and coverage $1-\mathrm{Abs}/n$, where $n$ is the total number of candidates. Candidates without a valid numerical claim remain in the evaluation and are labeled incorrect; Direct and the LLM judge abstain on them. All metrics are calculated from one fixed candidate set per benchmark, with full results reported in Table~\ref{tab:tA_appendix} (appendix). 
\FloatBarrier
\section{Results and Discussion}
\label{sec:results}
\begin{figure*}[t]
\centering
\includegraphics[width=0.95\textwidth]{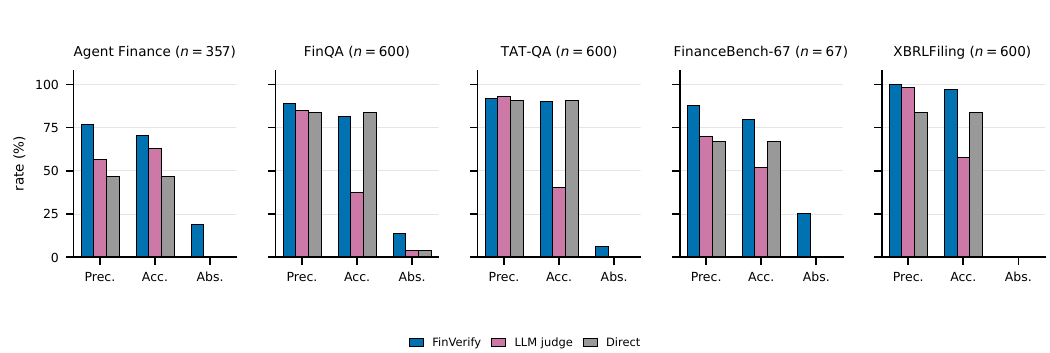}
\caption{Verification performance on the five benchmarks using fixed candidate sets. Bars
show accepted-claim precision, decision accuracy, and abstention rate per
verifier. Full counts are in
Table~\ref{tab:tA_appendix}.}
\label{fig:effectiveness}
\end{figure*}

All three verifiers judged the same candidates, with the inputs described in
Section~\ref{sec:verifiers}. Figure~\ref{fig:effectiveness} shows the five benchmarks.
Table~\ref{tab:tA_appendix} (appendix) lists every count.

\subsection{Verification performance}

\paragraph{Agent Finance Reasoning.}
The Agent Finance Reasoning benchmark contains 357 multi-step answers from five models, of which 190 are incorrect. Direct accepts all 190 incorrect answers, achieving 46.8\% precision, while the LLM judge accepts 120, with 56.4\% precision. FinVerify reduces false accepts to 21, rejects 135 incorrect answers, and abstains on 67 candidates, achieving 76.7\% precision and 81.2\% coverage. However, it also rejects 65 correct answers, compared with 12 for the LLM judge. Despite these false rejections, FinVerify achieves higher decision accuracy on non-abstained candidates (70.3\% versus 62.9\%).

\paragraph{FinanceBench-67 and XBRLFiling.}
For FinanceBench-67 and XBRLFiling, FinVerify checks candidate answers against the filed XBRL facts and calculation relationships in the underlying SEC filings. On XBRLFiling, it rejects 94 of 96 incorrect numerical answers, accepts one, and abstains on one. It also rejects 17 correct answers, achieving 99.8\% precision, 97.0\% decision accuracy, and 99.3\% coverage. On FinanceBench-67, where many questions require definitions or interpretations rather than reported financial values, FinVerify rejects 11 of 22 incorrect answers, accepts four, and abstains on seven, achieving 87.9\% precision, 80.0\% decision accuracy, and 74.6\% coverage. The four false accepts on FinanceBench-67 highlight a limitation of its reference answers. Each is written as a sentence, and the candidate's value either appears directly in that sentence (e.g., 20.0 versus ``approximately \$20 billion'') or represents the numerical change it describes (e.g., $-3.0$ for a decline from 24.6\% to 21.6\%). Neither the filed facts nor recomputation can establish a contradiction, and the LLM judge also accepts these four answers. In comparison, the LLM judge incorrectly rejects 247 correct answers on XBRLFiling and 22 on FinanceBench-67, resulting in decision accuracies of 58.0\% and 52.2\%, respectively.
\paragraph{FinQA and TAT-QA.}
For FinQA and TAT-QA, verification relies only on the provided text and tables, without access to filed XBRL facts or accounting identities. On FinQA, FinVerify achieves 89.0\% precision, compared with 83.9\% for Direct and 85.2\% for the LLM judge. Its decision accuracy is also higher at 81.6\%, compared with 37.7\% for the judge. On TAT-QA, FinVerify achieves 92.1\% precision, compared with 90.9\% for Direct and 93.1\% for the judge, while its decision accuracy reaches 90.0\% versus the judge's 40.7\%. FinVerify detects relatively few incorrect answers on these benchmarks, rejecting 25 on FinQA and four on TAT-QA. Its main advantage over the LLM judge is fewer false rejections: 46 versus 333 on FinQA and 13 versus 338 on TAT-QA. However, this comes with more abstentions: 83 versus 24 on FinQA and 39 versus five on TAT-QA.

\subsection{Discussion}
The results show that FinVerify's performance depends on the availability of authoritative financial evidence. On XBRLFiling, filed XBRL facts and accounting identities allow FinVerify to independently recompute reported values, resulting in high precision, decision accuracy, and coverage. On FinanceBench-67, many questions involve financial definitions or interpretations that cannot be verified directly against filed values. FinVerify can check the definitions stated in the question and the candidate's calculations, but an internally consistent answer based on the wrong financial item may still pass. This limitation is reflected in the four false accepts, where the reference answers are expressed in text and do not provide enough information to establish a contradiction. The other benchmarks show different trade-offs. On Agent Finance Reasoning, FinVerify accepts fewer incorrect answers than the LLM judge but also rejects more correct answers and abstains more often, increasing the need for human review. On FinQA and TAT-QA, where evidence is limited to text and tables, FinVerify achieves precision similar to that of the judge. However, the judge tends to reject correct answers, while FinVerify more often abstains when sufficient evidence is unavailable.
\section{Limitations}
\label{sec:limitations}
\paragraph{Generality of the taxonomy.}
The five error categories are derived from the 190 incorrect Agent Finance Reasoning trajectories that we also evaluate on, each classified by a single LLM call without human validation, so the result on this benchmark is in-sample. Testing whether the taxonomy generalizes requires held-out questions, other agent models, and answers from more than one generator; four of our candidate sets come from a single model.

\paragraph{Contribution of individual protocols.}
The source-fidelity and accounting protocols build on the evidence-grounded approach of VeriFin~\cite{verifin}; on FinanceBench-67 and XBRLFiling, FinVerify retrieves the SEC filings independently. We neither compare against VeriFin nor ablate the temporal, comparability, and financial-model protocols, and we do not report per-protocol results or analyze false accepts and false rejects in detail, so the contribution of each protocol remains unclear.

\paragraph{Baselines and coverage.}
The LLM judge uses a fixed zero-shot prompt and does not have access to all the evidence available to FinVerify. Specifically, it cannot access the underlying filing tables for Agent Finance Reasoning or the full SEC filings for FinanceBench-67 and XBRLFiling. A comparison with a judge given the same evidence and tools would provide a stronger evaluation. FinVerify also abstains more frequently than the baselines, so a risk--coverage analysis is needed for comparison at equal coverage. FinVerify is intended as an aid to human review, and practical deployment would require further testing on unseen agents and datasets.

\section{Conclusion}
\label{sec:conclusion}
We organized 190 incorrect financial-agent trajectories into five checkable
error categories; FinVerify turns them into evidence-grounded protocols in
which a language-model stage structures the inputs and deterministic checks
accept, reject, or abstain. It raises accepted-claim precision over Direct
on all five benchmarks and over an LLM judge on four: on multi-step agent
answers from 56.4\% to 76.7\% at 81.2\% coverage, with one false accept in
596 non-abstaining XBRLFiling decisions. On text-and-table QA the
differences are small and come with reduced coverage.

\paragraph{Future work.}
Future work will study how FinVerify's evidence bindings and protocol traces can support independently checkable certificates within the PCE architecture~\cite{certifiedtraces}. This requires binding certificates to claims, evidence snapshots, policy versions, and proposed actions; checking authorization and trace-level policy obligations beyond numerical correctness; and rejecting or escalating uncertified actions. Evaluation should measure unsafe authorizations and certification coverage alongside claim-verification accuracy.

\section*{Acknowledgments}
Xiao-Yang Liu acknowledges support from Columbia University's School of Engineering and Applied Science (SEAS) through the SEAS Interdisciplinary Research (SIRS) and SEAS Translational Acceleration Research (STAR) Seed Funding Program; the Tang Family Fund for Research Innovations in FinTech, Engineering, and Business Operations; the Center for Digital Finance and Technologies (CDFT); and the JPMorganChase Faculty Research Award.

The opinions expressed in this paper are those of the authors and do not necessarily reflect those of JPMorganChase or its affiliates. This material is not a product of the Research Department of J.P. Morgan Securities LLC and does not constitute investment advice, a recommendation, or a solicitation or offer in any jurisdiction.


\clearpage
\appendix
\section*{Benchmarks and full results}
Table~\ref{tab:pools} summarizes the five benchmarks of Section~\ref{sec:pools},
Table~\ref{tab:related} maps the error categories of Section~\ref{sec:taxonomy}
onto the prior work of Section~\ref{sec:related}, and Table~\ref{tab:tA_appendix}
lists the complete confusion counts behind Figure~\ref{fig:effectiveness} and
Section~\ref{sec:results}.
\begin{table}[t]
\centering
\footnotesize
\caption{The five benchmarks and their fixed candidate sets. Every verifier judges the same candidates and the same frozen candidate evidence; sources after ``+'' are additional authoritative material read by FinVerify only, not by the baselines (Section~\ref{sec:verifiers}). ``Wrong'' is the number of candidates that fail the correctness rule, and ``null'' the number with no parseable numeric claim (these stay in the candidate set).}
\label{tab:pools}
\begin{tabular}{@{}lllrrr@{}}
\toprule
Benchmark & Candidate generator & Evidence (+ FinVerify-only) & $n$ & wrong & null \\
\midrule
Agent Finance Reasoning~\cite{snorkel} & five backbones (dataset) & trace; + 10-K tables & 357 & 190 & -- \\
FinQA (dev) & Haiku 4.5 & FinQA text and table & 600 & 117 & 24 \\
TAT-QA (dev, arith.) & Haiku 4.5 & TAT-QA table and paragraphs & 600 & 59 & 5 \\
FinanceBench-67 & Haiku 4.5 & filing passages; + SEC filing & 67 & 22 & 0 \\
XBRLFiling & Haiku 4.5 & redacted excerpt; + SEC filing & 600 & 99 & 3 \\
\bottomrule
\end{tabular}
\end{table}

\begin{table}[t]
\centering
\footnotesize
\caption{Which error categories of Section~\ref{sec:taxonomy} each line of prior work checks. \ding{51}: checked against independent evidence; (\ding{51}): checked textually or on one evidence type, without the full typed context; --: not checked.}
\label{tab:related}
\begin{tabular}{@{}lccccc@{}}
\toprule
 & A Temporal & B Source & C Compar. & D Accounting & E Model \\
\midrule
Atomic-claim verifiers~\cite{factscore,factool,safe} & -- & (\ding{51}) & -- & -- & -- \\
LLM-as-judge~\cite{judges} & -- & (\ding{51}) & -- & -- & (\ding{51}) \\
Look-ahead benchmarks~\cite{leakage,lookahead,exante,guan2026leakage} & \ding{51} & -- & -- & -- & -- \\
FinGround~\cite{finground} & (\ding{51}) & \ding{51} & (\ding{51}) & -- & (\ding{51}) \\
VeriFin~\cite{verifin} & -- & \ding{51} & (\ding{51}) & \ding{51} & (\ding{51}) \\
\textbf{FinVerify (this work)} & \ding{51} & \ding{51} & \ding{51} & \ding{51} & \ding{51} \\
\bottomrule
\end{tabular}
\end{table}

\begin{table*}[t]
\centering\small
\caption{Full gate results on the fixed candidate sets of the five benchmarks. TA: correct answer accepted; FA: wrong answer accepted; TR: wrong answer rejected; FR: correct answer rejected; Abs: abstention. Precision is TA/(TA+FA), accuracy is (TA+TR)/(TA+FA+TR+FR), coverage is $1-\mathrm{Abs}/n$.}
\label{tab:tA_appendix}
\resizebox{\linewidth}{!}{%
\begin{tabular}{@{}llrrrrrlll@{}}
\toprule
Benchmark & Verifier & TA & FA & TR & FR & Abs & Precision \% & Accuracy \% & Cov.\ \% \\
\midrule
\multicolumn{10}{l}{\textbf{Agent Finance Reasoning ($n$=357, wrong=190)}} \\
 & FinVerify & 69 & 21 & 135 & 65 & 67 & 76.7 & 70.3 & 81.2 \\
 & LLM judge & 155 & 120 & 69 & 12 & 1 & 56.4 & 62.9 & 99.7 \\
 & Direct & 167 & 190 & 0 & 0 & 0 & 46.8 & 46.8 & 100.0 \\
\midrule
\multicolumn{10}{l}{\textbf{FinQA ($n$=600, wrong=117, 24 null)}} \\
 & FinVerify & 397 & 49 & 25 & 46 & 83 & 89.0 & 81.6 & 86.2 \\
 & LLM judge & 150 & 26 & 67 & 333 & 24 & 85.2 & 37.7 & 96.0 \\
 & Direct & 483 & 93 & 0 & 0 & 24 & 83.9 & 83.9 & 96.0 \\
\midrule
\multicolumn{10}{l}{\textbf{TAT-QA ($n$=600, wrong=59, 5 null)}} \\
 & FinVerify & 501 & 43 & 4 & 13 & 39 & 92.1 & 90.0 & 93.5 \\
 & LLM judge & 203 & 15 & 39 & 338 & 5 & 93.1 & 40.7 & 99.2 \\
 & Direct & 541 & 54 & 0 & 0 & 5 & 90.9 & 90.9 & 99.2 \\
\midrule
\multicolumn{10}{l}{\textbf{FinanceBench-67 ($n$=67, wrong=22)}} \\
 & FinVerify & 29 & 4 & 11 & 6 & 17 & 87.9 & 80.0 & 74.6 \\
 & LLM judge & 23 & 10 & 12 & 22 & 0 & 69.7 & 52.2 & 100.0 \\
 & Direct & 45 & 22 & 0 & 0 & 0 & 67.2 & 67.2 & 100.0 \\
\midrule
\multicolumn{10}{l}{\textbf{XBRLFiling ($n$=600, wrong=99, 3 null)}} \\
 & FinVerify & 484 & 1 & 94 & 17 & 4 & 99.8 & 97.0 & 99.3 \\
 & LLM judge & 254 & 4 & 92 & 247 & 3 & 98.4 & 58.0 & 99.5 \\
 & Direct & 501 & 96 & 0 & 0 & 3 & 83.9 & 83.9 & 99.5 \\
\bottomrule
\end{tabular}%
}
\end{table*}

\section*{Ethics}
An incomplete verifier in a financial setting risks \emph{false assurance}: an accepted claim is not an audit opinion. FinVerify is an advisory monitor whose localized flags go to a human reviewer, and it does not approve outputs. We report abstentions and null candidates so that coverage is not overstated. With human review, FinVerify can help catch unsupported financial claims and make numerical errors easier to investigate.

\section*{Reproducibility}
We store candidate sets, validated fits, verifier decisions, and language-model responses in JavaScript Object Notation Lines (JSONL) files indexed by task identifier, with prompts defined in code. All language-model calls use Claude Haiku~4.5 and are cached by role, task identifier, and prompt hash. Scores, figures, and tables can be regenerated from these records. For FinanceBench-67 and XBRLFiling, we use the candidate answers and correctness labels released with VeriFin~\cite{verifin}, but not its formulas or verification results. Instead, we retrieve and checksum fixed snapshots of XBRL facts and calculation linkbases from SEC EDGAR. Each candidate record also contains the evidence provided to the verifiers. Third-party assets are used under their published terms: FinQA (MIT), TAT-QA (Creative Commons Attribution, CC~BY~4.0), FinanceBench (Creative Commons Attribution-NonCommercial, CC~BY-NC~4.0), Agent Finance Reasoning (Apache~2.0), and Z3 (MIT); the XBRLFiling questions are used as distributed with the VeriFin artifact~\cite{verifin}, whose annotation license is pending, over public SEC filings.

Because the fitting prompt is independent of the protocol catalog, the comparability and financial-model policies in Section~\ref{sec:protocols} were evaluated on the same stored fits. We reran the deterministic protocol checks and proof gate using the stored fits, without regenerating candidates or making additional language-model calls. This isolates the effect of the added policies from variation in language-model outputs.

\section*{Z3 in FinVerify: encoding and a worked example}
\label{app:z3}

\paragraph{Encoding.}
FinVerify uses Z3~\cite{z3} in the authority engine shared by protocols D and E (Section~\ref{sec:protocols}). The engine selects a formula $f$ independently of the candidate's program and stated value. It searches three sources in order: a formula inferred from the question's wording (e.g., ``percentage change'' maps to $(\mathit{new}-\mathit{old})/\mathit{old}$), the issuer's calculation linkbase, and the metric registry. Explicit definitions stated in the question are handled by a separate policy without Z3.

Each operand $x_i$ is bound to a value $b_i$. Linkbase operands use filed XBRL facts, while question-derived or registry formulas use candidate operands. Protocol B checks their evidence grounding separately. Z3 runs only when all operands can be bound, regardless of whether B passes.

The engine checks satisfiability over the reals:
\[
\Phi=\bigwedge_i(x_i=b_i)\;\wedge\;(\hat v=f(x))
\;\wedge\;
\bigvee_{s\in\{1,\,1/100,\,100\}}
\left(|\hat v-sv|\le\tau\right),
\]
where $v$ is the stated value and $\tau=10^{-4}\max(|\hat v|,|v|,1)$. The scaling factors allow percentages and equivalent ratios to match. Each constraint is asserted under a separate name. A \textsc{sat} result passes the policy; \textsc{unsat} indicates that the stated value conflicts with the authoritative formula and bound operands. Such a conflict proves a violation only when operands are identified by XBRL concepts or fiscal periods. Bindings based solely on operand order are insufficient for rejection.

\paragraph{Worked example (XBRLFiling, \texttt{mc\_calc\_00191}).}
The question asks for Texas Instruments' FY2025 total current assets in USD millions. The candidate states 8,946, and the fitter reconstructs $3{,}225+1{,}656+1{,}963+2{,}102=8{,}946$ from the filing excerpt. All other applicable checks pass: source grounding (B), temporal integrity (A), comparability (C), and program consistency (D).

The issuer's calculation linkbase, however, defines total current assets using five filed values:
\[
\begin{aligned}
x_{\mathrm{Cash}}&=3{,}225, &
x_{\mathrm{STI}}&=1{,}656,\\
x_{\mathrm{AR}}&=1{,}963, &
x_{\mathrm{Inv}}&=4{,}804,\\
x_{\mathrm{Pre}}&=2{,}102.
\end{aligned}
\]
The authoritative formula gives
\[
\hat v=x_{\mathrm{Cash}}+x_{\mathrm{STI}}+
x_{\mathrm{AR}}+x_{\mathrm{Inv}}+x_{\mathrm{Pre}}
=13{,}750.
\]

Z3 receives seven named assertions: five operand bindings, the formula, and the claim constraint. With $\tau=1.375$, none of the three permitted scalings reconciles 8,946 with 13,750. Z3 returns \textsc{unsat}, and protocol D rejects the claim. The unsat core contains all seven assertions. The difference, $13{,}750-8{,}946=4{,}804$, is the inventory value omitted from the fitter's reconstruction. The benchmark's reference answer is \$13,750 million.
\clearpage
\end{document}